\documentclass[12pt]{elsarticle}
\usepackage{amsmath} 
\usepackage{amsfonts}
\usepackage{booktabs}
\usepackage{graphicx}
\usepackage[table]{xcolor}
\usepackage{cite}
 
\begin{document}
\begin{frontmatter}
\title{Discontinuous Piecewise Polynomial Neural Networks}
\author{John Loverich}
\address{N Infinity Computational Sciences\\
638 Beauprez Avenue\\
Lafayette CO 80026}
\begin{abstract}%
An artificial neural network is presented based on the idea of connections between units that are only active for a specific range of input values and zero outside that range (and so are not evaluated outside the active range).  The connection function is represented by a polynomial with compact support.  The finite range of activation allows for great activation sparsity in the network and means that theoretically you are able to add computational power to the network without increasing the computational time required to evaluate the network for a given input.  The polynomial order ranges from first to fifth order.  Unit dropout is used for regularization and a parameter free weight update is used.  Better performance is obtained by moving from piecewise linear connections to piecewise quadratic, even better performance can be obtained by moving to higher order polynomials.  The algorithm is tested on the MAGIC Gamma ray data set as well as the MNIST data set.
\end{abstract}


\end{frontmatter}

\section{Introduction}
Neural networks and their application to deep learning have become a focus of research due to the success of the algorithms on several types of problems \citep{Mnih2015,Bengio2009,Hinton2006}.  The goal of this work is to create a multi-layer artificial neural network where connections between units are active only in a finite range.  This naturally leads to grouping connections together into functional links that span the full range of input values.  Each connection then is called a sub link of the complete link.  Each sub link is represented using piecewise polynomial with compact support.  The resulting link looks like a one dimensional grid of discontinuous piecewise polynomials.  

The motivation for this work comes from the (1) the knowledge that activation of functional elements in biological neural networks is extremely sparse; (2) the idea that fewer layers in deep neural networks are required if one increases the computational capability of each layer; (3) the fact that the use of piecewise quadratic or higher order polynomials in an ANN results in increased polynomial output whereas the output of a piecewise linear network remains piecewise linear regardless of the number of layers; (4) increasing the sparsity (adding additional sub links) can theoretically increases computational power without increasing computational time; (5) higher order approximations are known to reduce the problem of adversarial examples \citep{Fawzi2015} and (6) it has been shown that dendrites (dendritic spines) are not just passive elements, but perform computations themselves \citep{Wilson2016}. 

The algorithm starts with a simple technique for approximating functions, using piecewise discontinuous polynomials.  Typically, discontinuous functions are avoided in artificial neural networks because gradient descent derivatives are not defined at the discontinuities.  In this paper, gradient descent is used as we wait for a alternate algorithms to be developed, one possible approach is recursive decomposition as described by Friesen in \citep{Friesen2015}.  The discontinuities act to break up the network into a series of sub networks.  Gradient descent is applied to each of these sub networks separately and good results are achieved.  The fact that this works is not entirely surprising given that the dropout technique \citep{Hinton2012,Srivastava2014} has the same effect on a network during training and is equivalent to training on a discontinuous network.  To see this, evaluate a network with half the hidden neurons dropped out and record the result.  Then select another set of random neurons to drop out and again evaluate the network with the same input.  The output from the two cases will be different and so the network is discontinuous during training as a result of dropout.  Note that at least one discontinuity per link should be used to remain a universal approximator as described by \citep{Leshno1993}. The discontinuous network tends to produce over fitted results in many problems. Over-fitting is resolved by using the dropout regularization technique described by \citep{Hinton2012,Srivastava2014}.  In addition, a parameter free weight update method as described by \citep{Schaul2012} is used to reduce the parameter search space.  Lagrange polynomials are used to describe the piecewise polynomial functions and, as such, the weights of the polynomial are the actual value of the polynomial at specific locations.  

In this paper, the link is the important computational element, but this approach can just as well be applied to the unit.  Piecewise polynomial approaches have been investigated in the CMAC architecture \citep{Lane1992} and more recently the rectified linear unit \citep{Nair2010}, has become a popular activation function which is piecewise linear .  High order neural network using higher order weighting terms are described by several authors including \citep{Giles1987,Shin1992,Elm2004,Foresti2004,Fallahnezhad2011} and functional link artificial neural networks (FLANN) by \citep{Pao1989,Patra2002,Purwar2007}, which is the approach used in this paper.   Discontinuous neural networks have been discussed in many articles, especially with respect to recurrent neural network including \citep{Forti2003,Liu2010,Gavalda1999} focused on convergence state estimation and stability and by \citep{Zhang2002} where a unique recurrent high order algorithm is derived for financial modeling, but where additional free parameters (weights) are added to the unit and a few simulations are performed with piecewise high order elements.  There is very little work on multi-layer high order discontinuous polynomial networks.  In this paper the algorithm is tested by performing simple curve fitting through the sine wave using a single link, and then classification with the MAGIC gamma ray detection data set and the MNIST data set.  The MNIST test is performed using multiple autoencoders.  The unique contribution of this paper is the application and development of novel algorithm, using discontinuous piecewise polynomial approximation, in a multi-layer neural network as well as the use of the parameter free weight update described by \citep{Schaul2012}.  The algorithm opens the possibility of using a variety of complicated, discontinuous elements in an artificial neural network using back propagation.

Section \ref{S:Algorithm} describes the algorithm used in this paper including backpropagation with discontinuities, weight initialization and link input range selection.  In Section \ref{S:Results} we demonstrate the algorithm on 3 problems: a simple 1D function approximation; the Magic gamma ray data set; the MNIST data set. In Section \ref{S:Conclusion} we conclude the paper.

\section{Algorithm}\label{S:Algorithm}

Definitions used in this paper are defined in table \ref{T:Definitions}.
\begin{table}[htb]
\begin{tabular}{l l}
Link & Is the connection between two units.\\
Sub Link & A link is split into sub links.\\
$\omega_{i}$ & The $i^{th}$ weight of a sub link.\\
$\omega_{j\,,\alpha}$ & the $\alpha^{th}$ weight of the $j^{th}$ link.\\
$B_{i}$ & The $i^{th}$ basis function of a sub link.\\
$N_{in}$ & The number of active links into a unit.\\
$N_{out}$ & The number of active links out of a unit.\\
$N_{p}$ & The number of Chebyshev-Lobatto nodes.\\
$x_{i}$ & The input to the $i^{th}$ link.\\
$x_{j}$ & The $j^{th}$ Chebyshev-Lobatto node.\\
$y_{m\,i}$ & The measured output value for output link $i$\\
$y_{d\,i}$ & The desired output value for output link $i$\\
$F_{i}$ & The output function for link $i$.\\
$y_{i}$ & The output for link $i$.\\
$E$ & The network error for a single input.\\
$t_{N_{p}}$ & Time to completion using sub links with $N_{p}$ nodes
\end{tabular}\caption{Variable definitions used in this paper.}\label{T:Definitions}
\end{table}
There are two natural ways to apply higher order approximations in artificial neural networks.  The first is to replace the single weight at the link with multiple adjustable weights describing a more complicated link function - this is a functional link neural network (FLANN) as described by \citep{Pao1989} and demonstrated by \citep{Patra1999}.  The alternative is to add adjustable parameters to the unit that describe a changing activation function, in this case adjustable parameters exist in both the unit and the link.  In this paper we chose the FLANN approach as it can be written slightly more compactly while maintaining weights defined at the link.

\begin{figure}[hb]
\begin{center}
\includegraphics[scale=0.5]{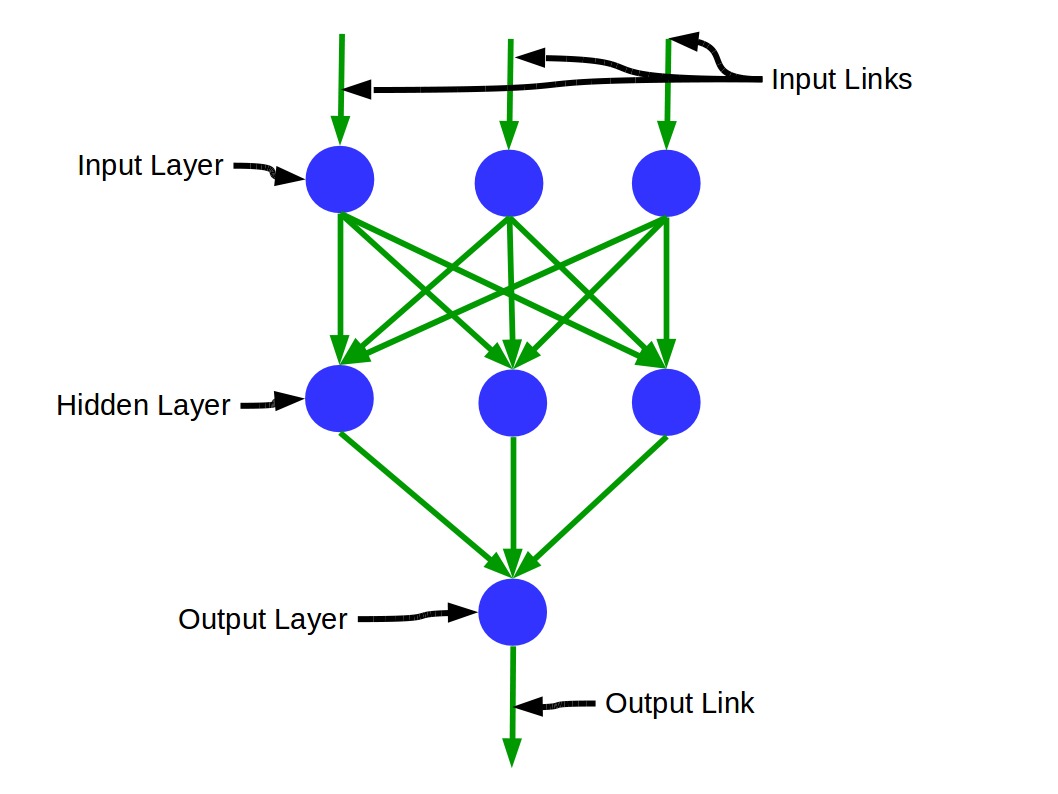}
\caption{A standard feed forward network.  Blue circles represent summing unit and green lines represent links where the signal passes from one neuron to the next along the direction of the arrow.}\label{F:FeedForward}
\end{center}
\end{figure}

\begin{figure}[hb]
\begin{center}
\includegraphics[scale=0.5]{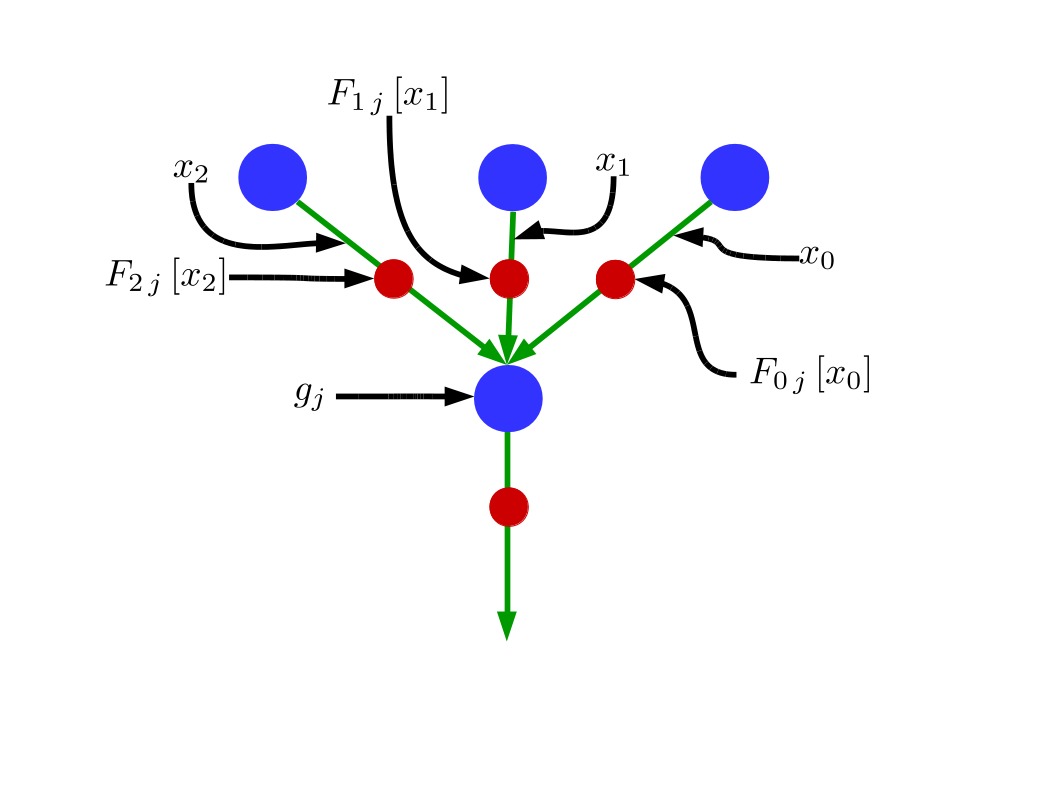}
\caption{Zoom in of connected units.}\label{F:NetworkZoom}
\end{center}
\end{figure}

The error correction algorithm used is backpropagation as described by \citep{Rumelhart1988} applied to a FLANN with a minor modification described in Section \ref{S:Backprop}.  The weight update rule is defined by using the parameter free weight update rule described by \citep{Schaul2012}.  A slight modification to this rule is that the maximum learning rate is set to 0.9.  The network description is that of a standard feed forward network, see Figure \ref{F:FeedForward}, with input links and output links added.  Labels for a network element are shown in figure \ref{F:NetworkZoom}.  The main differences of the algorithm compared to a standard network \citep{Rumelhart1988} are: (1) there are multiple weights per link; (2) No bias units are used; (3) the unit averages the input signal to produce an output instead of applying a more complex activation function; (4) input/output links are added which can be used to normalize and shift the data to the desired input/output range. 

The weight function of a sub link is described by the following equation,
\begin{equation}
f\left(x\right)=\sum_{i}w_{i}B_{i}\left(x\right)\,,
\end{equation}\label{E:weightFunction}
with the basis functions given by the Lagrange polynomials
\begin{equation*}
B_{j}=\prod_{0\leq m\leq k,m\neq j}\frac{\left(x-x_{m}\right)}{\left(x_{j}-x_{m}\right)}\,.
\end{equation*}
The Lagrange polynomial $B_{j}$ is useful because it has the value $1$ at $x=x_{j}$ and has the value 0 at all other $x_{i}$.  The interpolation nodes are given by the Chebyshev-Lobatto nodes, these differ from pure Chebyshev nodes in that the end points of the domain are included.  The Chebyshev-Lobatto nodes are given by,
\begin{equation*}
x_{j}=-\text{cos}\left(\frac{k\,\pi}{N_{p}-1}\right)\,,
\end{equation*}
where $k$ ranges from $0$ to $N_{p}-1$ and $N_{p}$ is the total number of Chebyshev-Lobatto points. 
This means that in Equation \eqref{E:weightFunction} the value of the function at $x_{j}$ is $w_{j}$.  Using this approach we can easily limit the range of a polynomial interpolation by limiting the range of the weights $w_{i}$.  For clarity we provide the polynomials for both a linear and quadratic interpolation.  In the linear case (assuming $x_{0}=-1$ and $x_{1}=1$)
\begin{equation*}
B_{0}=-\frac{1}{2}\left(x-1\right)
\end{equation*}
\begin{equation*}
B_{1}=\frac{1}{2}\left(x+1\right)
\end{equation*}
In the quadratic case we assume $x_{0}=-1$, $x_{1}=0$, $x_{2}=1$ the basis functions are
\begin{equation*}
B_{0}=\frac{1}{2}x\left(x-1\right)
\end{equation*}
\begin{equation*}
B_{1}=-\left(x+1\right)\left(x-1\right)
\end{equation*}
\begin{equation*}
B_{2}=\frac{1}{2}\left(x+1\right)x
\end{equation*}

In addition to the high order weighting, discontinuities are used along with the piecewise polynomial approximation.  This means that the function is different depending on the range of the input variable $x$.  In particular we have the definition
\begin{equation}
	F_{i}\left(x_{i}\right)=\left(
	\begin{array}{cc}
	\text{if}\left[r_{min}\leq x < a_{0}\right] & f_{i\,0}\left(x_{i}\right) \\
	\text{if}\left[a_{0}\leq x < a_{1}\right] & f_{i\,1} \left(x_{i}\right)\\
	... & \\
	\text{if}\left[a_{n}\leq x \leq r_{max}\right] & f_{i\,n}\left(x_{i}\right) \\
	\end{array}
	\right)
\end{equation}\label{E:F}

At the unit, the average of the incoming signals is computed.  The unit could be a sigmoid, but it is not needed since the non-linearity is provided by the presence of at least one discontinuity, so only a simple average is used as the activation function of the unit.
\begin{equation*}
g=\frac{1}{N_{in}}\left[\sum_{j}^{N}F_{j}\left(x_{j}\right)\right]
\end{equation*}
The averaging is important because the function $F_{i}$ only has a valid solution in a finite range so it is important to guarantee that any signal passed to $F_{i}$ is within that range.  This can easily be accomplished by choosing initial $r_{min}$ and $r_{max}$ correctly.  $N_{in}$ is the number of input signals (not the total number of input sub links).

\begin{figure}
\begin{center}
\includegraphics[scale=0.3]{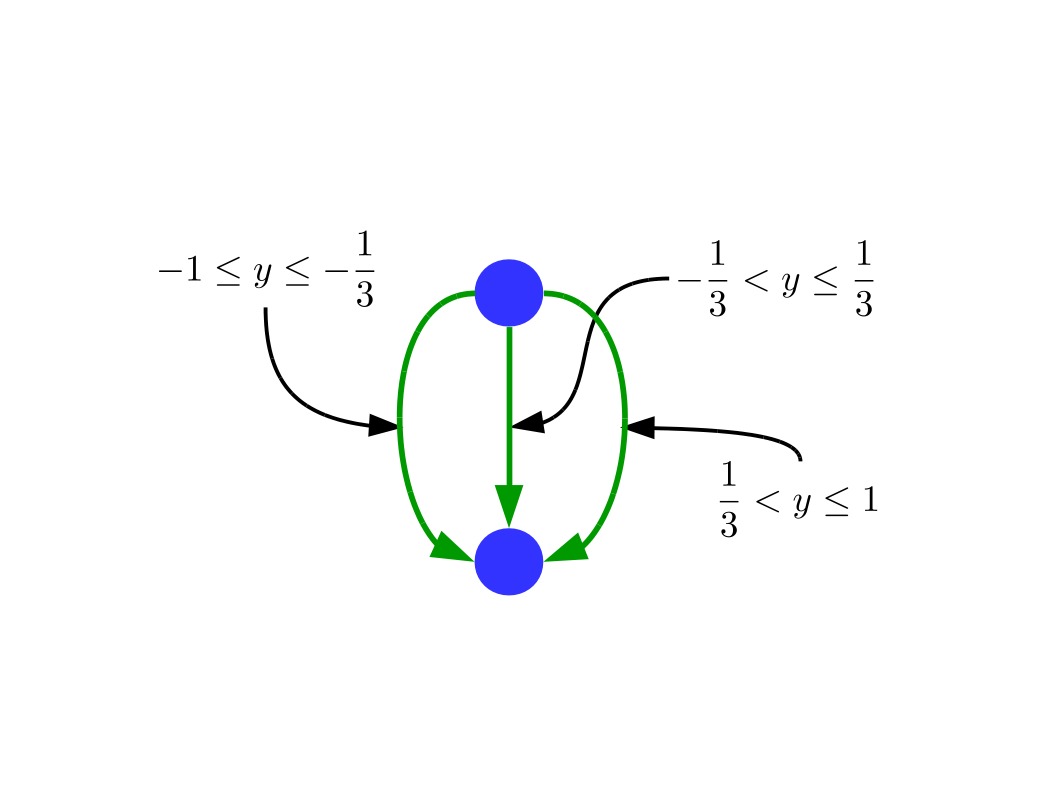}
\caption{Two units are connected by several links.  Each link is only active for a range of possible inputs.  This links can be combined into a single link with multiple weights and discontinuities between sub links.}\label{F:Pumpkin}
\end{center}
\end{figure}

\begin{figure}
\begin{center}
\includegraphics[scale=0.3]{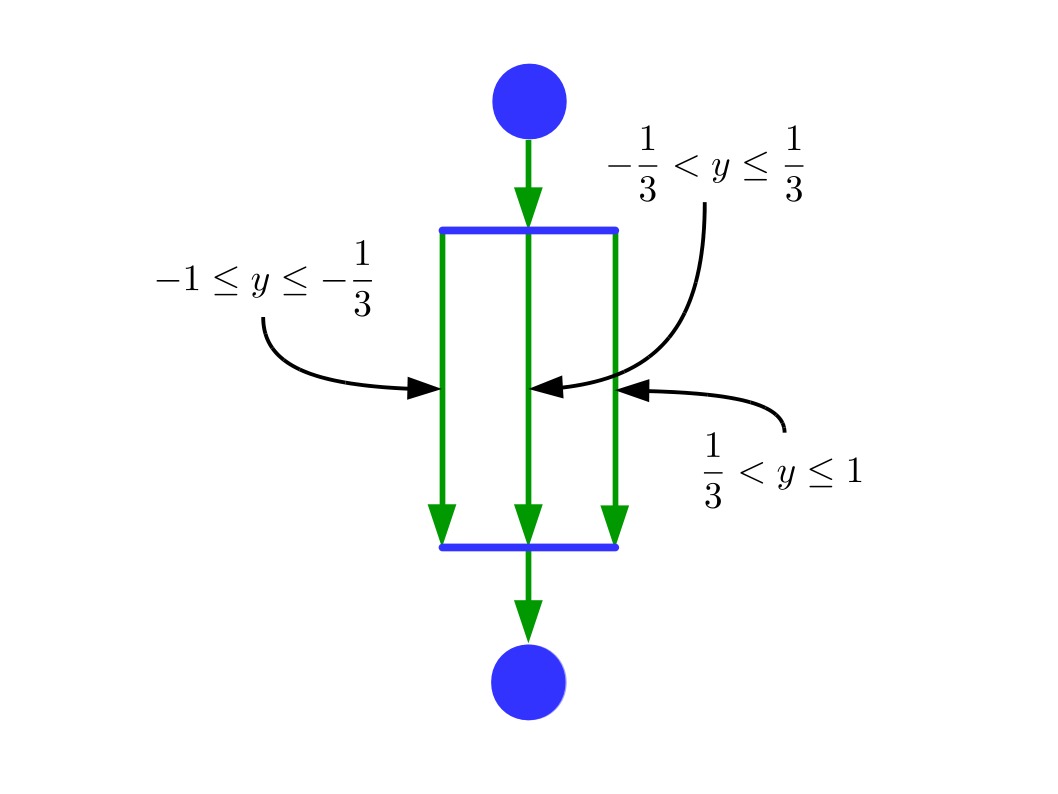}
\caption{A pair of units connected by a single link with 3 sub links where only one sub link is active for a given input signal.  This grouping of links is called a ``bundle".  This is the simplified approach taken in this paper.}\label{F:Bundle}
\end{center}
\end{figure}

\begin{figure}
\begin{center}
\includegraphics[scale=0.5]{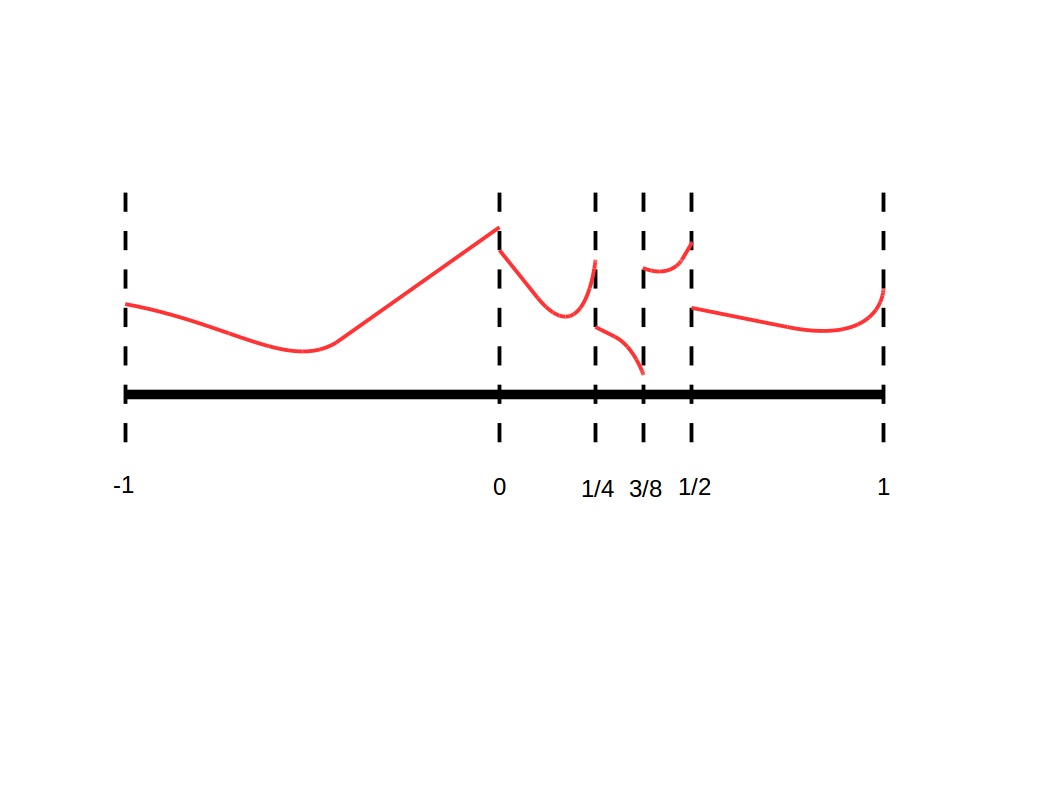}
\caption{A link can be represented as a grid with a piecewise
function valid on each of the ranges of the sub links.  The link function is discontinuous at the ends of each range.  Notice that in this link the sub link ranges are not equally spaced - and they don't need to be.  For the purposes of this paper, all ranges are equally spaced.}\label{F:amr}
\end{center}
\end{figure}

The idea to use a discontinuous piecewise polynomial comes from a pair of units with multiple links between the units, see Fig. \ref{F:Pumpkin}.  Only a single link is active depending on the output value of the unit.  Equation \eqref{E:F} can be described as a set of links between two units where only one link passes an output signal for a given input signal.  As a result, the links are grouped together in a bundle (Equation \ref{E:F}) and shown in figure \ref{F:Bundle}.  The standard link of a neural network described in this paper then consists of one or more sub links.  The link function can now be thought of as a one dimensional grid with piecewise polynomial elements within each grid cell as shown in Figure \ref{F:amr}.
\clearpage
\subsection{Backpropagation with Discontinuities}\label{S:Backprop}
\begin{figure}
\begin{center}
\includegraphics[scale=0.3]{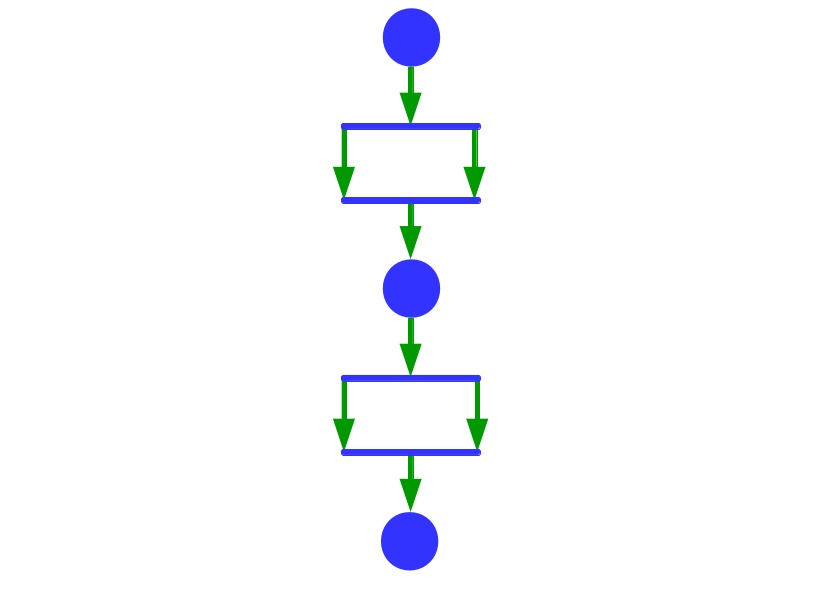}
\caption{Two stacked bundles and their associated units. In this case each bundle has two sub links and therefore a signal can take one of two paths for each bundle.}\label{F:Stacked}
\end{center}
\end{figure}

\begin{figure}
\begin{center}
\includegraphics[scale=0.6]{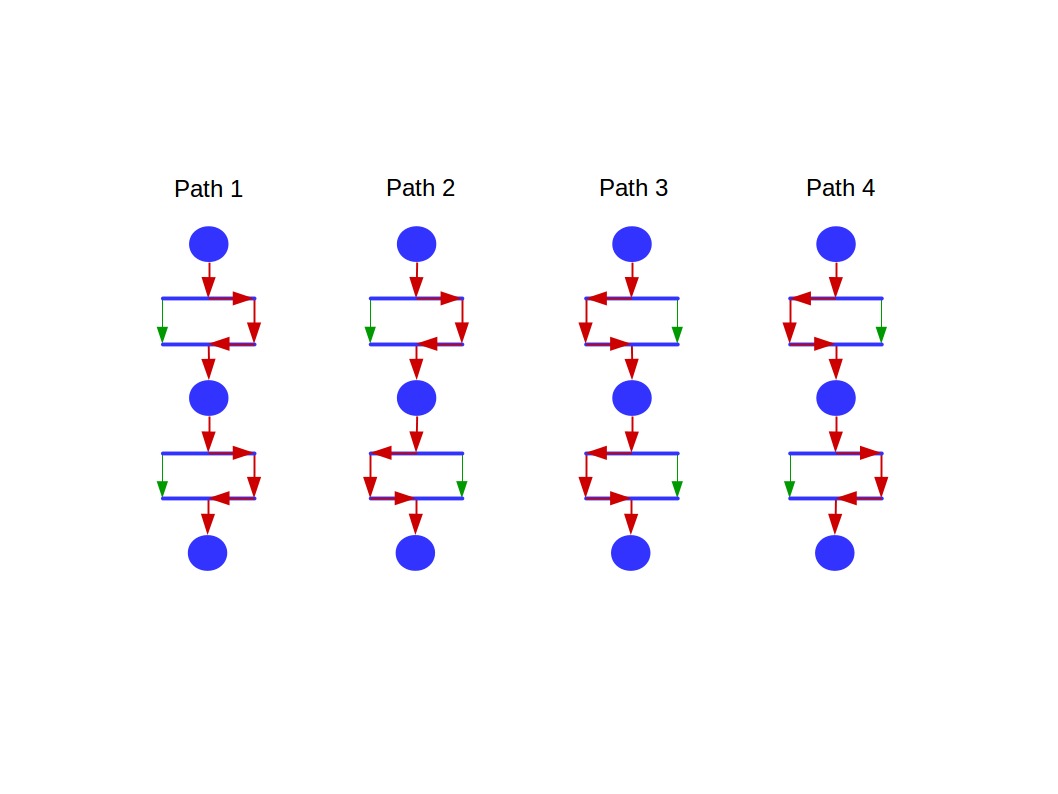}
\caption{The total number of paths for two bundles stacked on top of one another, with 2 sub links each, is 4.  The 4 paths are drawn out in this diagram.  Only the active path is used in the back propagation algorithm for each signal.  This means that the network is actually 4 separate coupled networks.  The coupling occurs because some of the weights are shared between networks (whenever part of a path is shared, the weights are shared).}
\label{F:paths}
\end{center}
\end{figure}

A network using a link with at least two sub links has the following property: for a given input signal only a subset of the network is active.  When training is performed, only the weights of the activated sub links are updated.  Figure \ref{F:Stacked} shows a 3-layer network consisting of two links, each link with two sub links.   Figure \ref{F:paths} shows this same network with all the possible paths that a signal can take from input to output.  It is evident from Figure \ref{F:paths} that there are 4 networks where each network shares some weights with the other networks.  It's easy to over train this type of network since each signal only effects a subset of the weights.  To resolve over training the ``dropout" regularization technique is used as described in \citep{Hinton2012}.

The key to getting an algorithm working with functions that are discontinuous is a backpropagation algorithm that works for discontinuous functions.  The implementation is incredibly simple and the only requirement is to remove idle sub links from the back propagation step.  After a signal has been propagated through the network, back propagation is performed, but only on the subset of the network active for the input signal.  This means that for each link, only one sub link is active, and so back propagation is performed only on that sub link, all other sub links in the link are ignored.  When weights are updated, they are only updated in the links that fired during the forward step.

For clarity we include a description of the back propagation applied to this network, recall that backpropagation is only applied to the active sub network.
\begin{itemize}
\item Forward propagate input signal
\item Record each sub link that is activated for the given input signal
\item Back propagate error signal through active sub links
\item Update weights of the active sub link
\item Ensure that weights are within desired range, 
$w_{i}=min\left(w_{max},max\left(w_{min},w_{i}\right)\right)$
\item Repeat with a new input signal
\end{itemize}

The error at the output of the network is measured as
\begin{equation*}
E=\sum_{i}^{N_{i}}\frac{1}{2}\left(y_{m,i}-y_{d,i}\right)^{2}
\end{equation*}
\begin{equation*}
\frac{\partial E}{\partial y_{m,i}}=y_{m,i}-y_{d,i}
\end{equation*}
The derivative of a link output with respect to a link input is given by
\begin{equation*}
\frac{\partial y_{i}}{\partial x_{i}}=\frac{\partial F_{i}}{\partial x_{i}}
\end{equation*}
Although, $F_{i}$ is discontinuous at points, it is entirely continuous within the range of the derivative.  If the derivative happens to be required exactly at the discontinuity, the derivative is taken only in the activated sub link (either left or right of the discontinuity).  The derivative across a unit to one of the unit input links is
\begin{equation*}
\frac{\partial y_{j}}{\partial x_{i}}=\frac{1}{N_{in}}
\end{equation*}
The derivative of the weight with respect to the link output is
\begin{equation*}
\frac{\partial y_{j}}{\partial w_{j,\alpha}}=\frac{\partial F_{j}}{\partial w_{j,\alpha}}
\end{equation*}
Error at the top of the link in the output link
\begin{equation*}
\delta E_{i}=\frac{\partial E}{\partial x_{i}}=\frac{\partial y_{i}}{\partial x_{i}}\frac{\partial E}{\partial y_{i}}=\frac{\partial y_{i}}{\partial x_{i}}\left(y_{i}-y_{d\,i}\right)
\end{equation*}
Error at the top of the link one layer in
\begin{equation*}
\delta E_{j}=\frac{\partial E}{\partial x_{j}}=\frac{\partial y_{j}}{\partial x_{j}}\frac{\partial E}{\partial y_{j}}=\frac{\partial y_{j}}{\partial x_{j}}\sum \frac{\partial x_{i}}{\partial y_{j}}\frac{\partial E}{\partial x_{i}}=\frac{\partial y_{j}}{\partial x_{j}}\sum \frac{\partial x_{i}}{\partial y_{j}}\delta E_{i}
\end{equation*}
In the specific case of an averaging unit, the error is
\begin{equation*}
\delta E_{j}=\frac{\partial y_{j}}{\partial x_{j}}\frac{1}{N_{in}}\sum \delta E_{i}
\end{equation*}
The rule one layer in can be applied to all further layers.  The error in the weight is then
\begin{equation*}
\frac{\partial E}{\partial w_{\alpha,j}}=\frac{\partial y_{j}}{\partial w_{\alpha,j}}\frac{\partial E}{\partial y_{j}}=\frac{\partial y_{j}}{\partial w_{\alpha,j}}\frac{1}{N_{in}}\sum \delta E_{i}
\end{equation*}
At this point, all weight update rules that work for standard neural networks will work for this algorithm as well.  A simple example is the momentum update
\begin{equation*}
w_{\alpha,j}^{n+1}=w_{\alpha,j}^{n}-\mu\frac{\partial E}{\partial w_{\alpha,j}}+\gamma\left(w_{\alpha,j}^{n}-w_{\alpha,j}^{n-1}\right)
\end{equation*}
Although the simple update works for this problem, we instead use Schaul's parameter free update \citep{Schaul2012}.

\subsection{Accelerated Backpropagation}
The backpropagation algorithm described is the standard algorithm applied to this model.  It was pointed out by \citep{Aizenberg2007} that a simple modification to the back propagation algorithm distributes the error more evenly among units.  In particular, the algorithm shown suggests that an error at the unit is $1/N_{in}$ times the error at the top of the output link (the links leaving the unit).  The error at the output of the unit is distributed evenly over the input links, and therefore the more input links there are, the smaller the error that each input link receives.  A simple way to remedy this problem is to replace the back propagation $1/N_{in}$ (one over the number of active links entering a unit) with $1/N_{out}$ (one over the number of links leaving a unit) which will result in a more even distribution of weight change throughout the network and faster convergence.  This technique is used in this paper.

%\begin{figure}[hb]
%\includegraphics[scale=0.5]{amr2}
%\caption{The same bundle where continuity is enforced between each of the ranges.  Continuity is enforced %after a range update.  Since only one range is active per update, the neighboring range nodes are simply set %to the nodal values of the range that fired.}\label{F:amr2}
%\end{figure}

\clearpage
\subsection{Weight Initialization}
Weight initialization is a huge concern for neural networks (\citep{Yam2000}).  In this paper, the weights within a link are initialized such that $F(x)$ (Equation \ref{E:F}) is a line across
the link, the equation is given as
\begin{equation}
\omega_{i}=\left(\frac{x_{i}-r_{\text{min}}}{r_{\text{max}}-r_{\text{min}}}\right)\omega_{\text{a}}+\omega_{\text{a}}\,,
\end{equation}\label{E:WeightInit}
where $w_{a}$ is chosen with a uniform random distribution in the range $[-1,1]$.
  
%As stated in the previous section, if the decay factor of a signal from one layer to the next is given by $w_{max}/r_{max}$ so it is best to choose the $w_{max}=r_{max}$ for interior bundles.  The input links and output link can have arbitrary ratios to rescale the input and output to a desired range.
\subsection{Choosing Ranges $r_{\text{min}}$ and $r_{\text{max}}$}\label{S:ChooseRange}
\begin{figure}
\begin{center}
\includegraphics[scale=0.75]{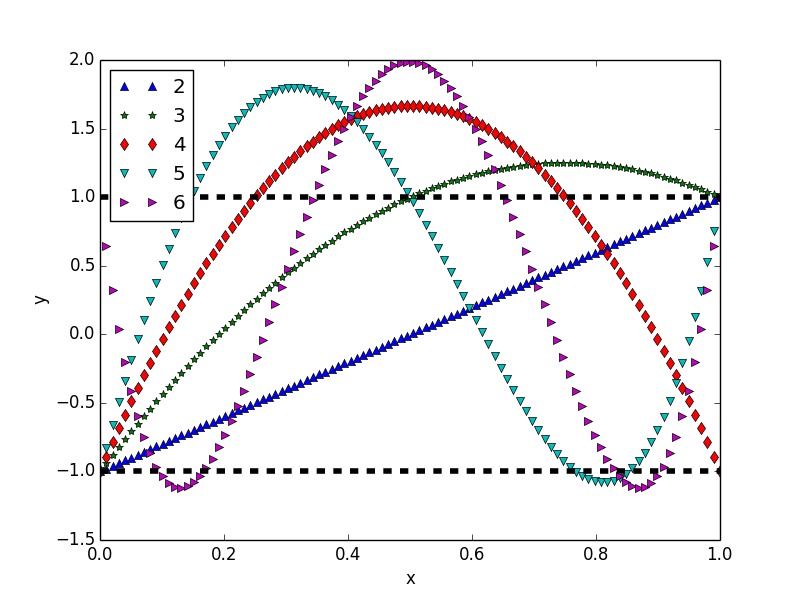}
\caption{Lagrange polynomials with weights in the range $\left[-1,1\right]$ chosen for maximum overshoot.  Even though the weights are limited to values in the range $\left[-1,1\right]$ the function can produce values outside this range.  The next link then should have an input range
that accounts for this overshoot.}\label{F:Lagrange}
\end{center}
\end{figure}

Choosing proper ranges for the links is key to getting good solutions.  The weights used in the Lagrange polynomial interpolation do not mark the limits of the polynomial value except in the linear case.  Figure \ref{F:Lagrange} shows 5 Lagrange polynomials with maximum overshoot for weights within the desired range - note that only in the linear case is the range limited by the values of the weights.  Instead, if the weights are in the range $[-\omega_{\text{max}},\omega_{\text{max}}]$ then a given choice of weights will produce a maximum overshoot $p_{n,\text{max}}\omega_{\text{max}}$ where values of $p_{n,\text{max}}$ are given in Table \ref{T:OverShoot}.  For a given input, the maximum possible output would be $p_{n,\text{max}}\omega_{\text{max}}$ which could then be the input to the next link.  The input range for each link should be set to $\left[-p_{n,\text{max}}\omega_{\text{max}},p_{n,\text{max}}\omega_{\text{max}}\right]$ to account for these overshoots.
One potential consequence of this choice of range is input value decay.  Consider a deep network with only one unit per layer and one link between units.  Each layer is initialized using Equation \eqref{E:WeightInit} with $w_{a}=1$ and input range $\left[-p_{n,\text{max}},p_{n,\text{max}}\right]$
Suppose the input value is $x_{\text{in}}$ then as the input value passes through each layer it will be compressed if $r_{max}=p_{n,\text{max}}>\omega_{max}$ by the ratio $\omega_{max}/r_{max}$.  After passing through $n$ layers, the output signal will be $\left(\omega_{max}/r_{max}\right)^{n}x_{in}$.  As a consequence, the output from each layer is pushed towards the value 0 which means fewer and fewer of the network sub links are used.  Fortunately, two things can occur to help the situation: (1) if a discontinuity is at the origin rapid learning can still occur since if a signal is either side of 0 it will adjust disconnected weights; (2) as the network is trained, more and more of the sub links are used.  It would seem randomly initializing the weights could alleviate this problem, however, we've found that symmetric initialization as in Equation \eqref{E:WeightInit} works better.  In this paper we use $r_{max}=-r_{min}=p_{n,\text{max}}$ where $p_{n,\text{max}}$ is provided in Table \ref{T:OverShoot} which gives the maximum overshoot for the polynomials when the weights are constrained to be between 1 and -1.
\begin{table}
\begin{center}
\begin{tabular}{ ccccccccccc }
\toprule
  number of points & 2 & 3 & 4 & 5 & 6 & 7 & 8 & 9 & 10\\
  \midrule
  $p_{n,\text{max}}$ & 1 & 1.25 & 1.667 & 1.799 & 1.989 & 2.083 & 2.203 & 2.275 & 2.362\\
  \bottomrule
\end{tabular}\caption{Maximum overshoot fraction, $p_{n,\text{max}}$, given the number of points in the polynomial interpolation (to 4 digits rounded up).}\label{T:OverShoot}
\end{center}
\end{table}
\clearpage
\subsection{Automatic Normalization}
Since input links and output links are used in this network and the range and weight is arbitrary, data
does not need to be normalized before passing to the network.  Normalization automatically occurs for input links by the user's choice of $r_{min}$ and $r_{max}$ for the input links, and by the users choice of $w_{min}$ and $w_{max}$ for the output links (see Figure \ref{F:FeedForward} for definitions of input and output links).  For example, if the MNIST data is being used and image values range from 0 to 255, simply set $r_{min}=0$ and $r_{max}=255$ in the input link.  Similarly, if the output values should be between 0 and 100 then set the weight limits in the output link to $w_{min}=0$ and $w_{max}=100$.  Although not particularly important, this is a convenient way to deal with input and output without a separate normalization step.
\section{Results}\label{S:Results}
In this section we use the network on 3 problems.  The first is the simple sine wave to illustrate how the functional link approximates this function. This provides clarity for how the link in this network differs from the single weight used in the standard approach.  The second problem uses the MAGIC gamma ray detection data set to predict if a gamma ray has been detected or not.  Finally, results for the MNIST problem are computed with 10 autoencoder classifiers (one for each digit) in a manner similar to that described in \citep{Kamyshanska2013}, though using the reconstruction error to determine the digit.  All results are computed using online backpropagation.  In this problem, all input and output links have fixed weights and the link function is linear with $\omega_{0}=-1$ and $\omega_{n}=1$.  Input links have a range $[r_{\text{min}},r_{\text{max}}]$ dependent on the range of the input parameters.  50\% dropout is used in all cases except the sin wave which only uses one link.
\subsection{Sine Wave}
\begin{figure}
\begin{center}
\includegraphics[scale=0.75]{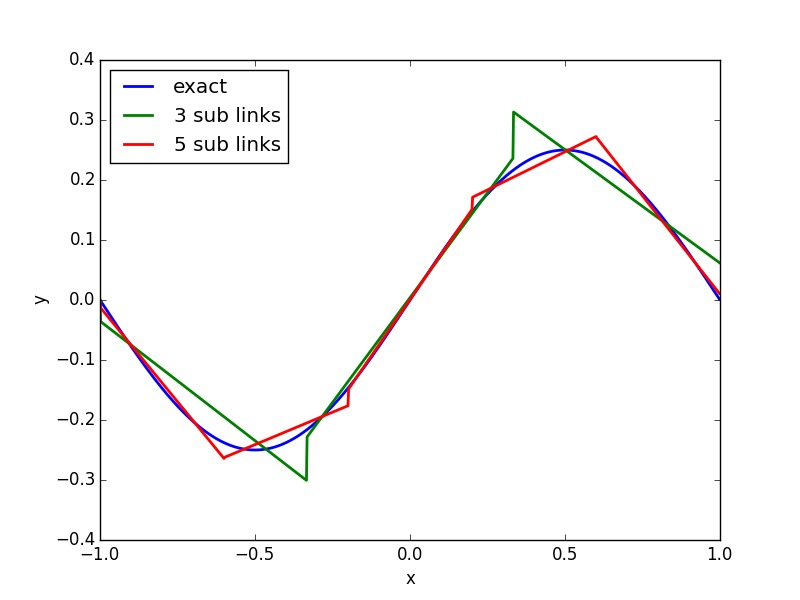}
\caption{A single link used to approximate a sine wave with piecewise linear sub links ($N_{p}=2$).  As the number of elements increases the fit improves.  Discontinuities are visible at sub link boundaries, the level of discontinuity decreases as the number of sub links is increased.}\label{F:DGSin1}
\end{center}
\end{figure}

\begin{figure}
\begin{center}
\includegraphics[scale=0.75]{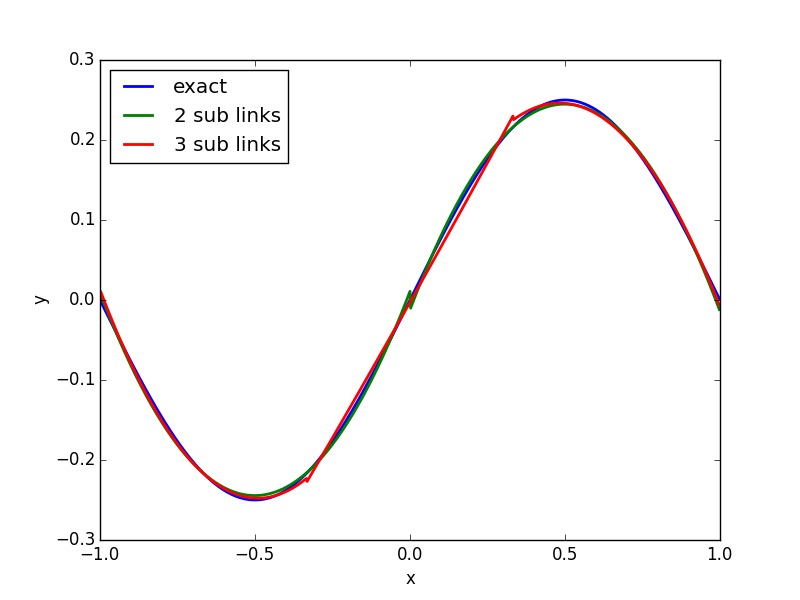}
\caption{A single link used to approximate a sine wave with piecewise quadratic elements ($N_{p}=3$).  In this case we get a much better match than the case with piecewise linear elements even with fewer total weights.  This is a well-known feature of higher order approximations.}\label{F:DGSin2}
\end{center}
\end{figure}

A sine wave is approximated using a single link with several sub links showing the function approximation capability of a single link.  This problem is illustrative of how the functions in each sub link are combined to produce the full function.  Figure \ref{F:DGSin1} shows a sine wave approximated using a bundle with 3 and 5 linear sub links where discontinuity is allowed between the sub links.  Figure \ref{F:DGSin2} shows the same sine wave approximated using a link with 2 and 3 quadratic sub links with discontinuity between them.  Note that the linear approximation has 6, and 10 degrees of freedom (for 3 and 5 ranges) and the quadratic approximation has 6 and 9 degrees of freedom (for 2 and 3 ranges).  Despite having fewer degrees of freedom the quadratic approximation is substantially better than the linear approximation.  This just illustrates the fact that higher order polynomial approximations require fewer degrees for freedom (for the same accuracy) than lower order approximations during function approximation.
\clearpage
\subsection{Gamma Ray Detection}
The gamma ray detection data is data created to simulate the detection of gamma rays in a ground based atmospheric Cerenkov gamma ray telescope.  There are 19,020 examples, where 2/3 of the data is used for training and validation and the remaining 1/3rd is used for testing.  The data is shuffled 10 times using a different set for training and test.  In the validation runs the 10\% of the training data is used as the validation set. 

The data can be obtained from the University of California Irvine repository \citep{Lichman2013}.  In the results that follow a parameters scan for network geometry [3, 5, 10, 20, 80, 160] neurons in the layer and with [1, 2, 3, 4, 5, 6, 7, 8] layers,
number of sub links was varied from 1 to 6 over one epoch and with $N_{p}=2$ to $6$.  The solution that produced the best validation accuracy was the case with 80 units in the hidden layers, and 2 sub links.
Using this network geometry we then ran $N_{p}=2$ to $6$ for 50 epochs, using 10 separate test and training sets.  The results are shown in Table \ref{T:GammaRay}.  Table \ref{T:GammaRay2} shows previously published results using a decision tree with softening splits\citep{Dvovrak2007} for comparison.  Table \ref{T:GammaRayConfidence} gives the confidence scores as a percent chance that the means of the $N_{p}$ in the columns and the means of the $N_{p}$ in the rows actually match for the results of table \ref{T:GammaRay}.

Table \ref{T:GammaRay} shows that piecewise linear $N_{p}=2$ does not perform nearly as well as the higher $N_{p}$ solutions. The performance on this problem reaches a minimum around $N_{p}=5$.  In particular, the performance improvement from $N_{p}=2$ to $N_{p}=3$ is about 25\%.  A key difference between the piecewise linear case $N_{p}=2$ and the higher order interpolations is that even after N layers are added to the network, the output of a piecewise linear network, is still piecewise linear, where as if an nth order polynomial is used (for $n>1$) as the link function, the output polynomial function is of order $n\,k$ where $k$ is the number of layers of links.
\begin{table}
\begin{center}
\begin{tabular}{ccc}
\toprule
  $N_{p}$ & test error & $\sigma$ \\
  \midrule
  2 & 0.2065 & 0.0225\\
  3 & 0.1474 & 0.0076\\
  4 & 0.1368 & 0.0053\\
  5 & 0.1311 & 0.0039\\
  6 & 0.1318 & 0.0041\\
\bottomrule
\end{tabular}\caption{Results on MAGIC data set based on the average of 10 splits of the original data set.  Each case
was run for 50 epochs on a fully connected network with
4 hidden layers of 50 neurons each, and 2 sub links per link. $\sigma$ is the standard deviation of the averaged results.}\label{T:GammaRay}
\end{center}
\end{table}

\begin{table}
\begin{center}
\begin{tabular}{ccccc}
\toprule
$N_{p}$ & 2 & 3 & 4 & 5\\
\midrule
3 & 1.0e-3 \\
4 & 3.1e-4 & 3.8e-1 \\
5 & 2.9e-4 & 6.2e-3 & 2.4 \\
6 & 3.2e-4 & 1.0e-2 & 4.9 & 100 \\
\bottomrule
\end{tabular}\caption{T-test table showing the \% risk that the means given in table \ref{T:GammaRay} are actually equal.  Note that all scores are less than 5\% except the case comparing 6 and 5.  What this table shows is that higher order accuracy produces significantly better results in this problem until $N_{p}=6$ where the difference is insignificant.}
\label{T:GammaRayConfidence}
\end{center}
\end{table}

\begin{table}
\begin{center}
\begin{tabular}{ cc }
\toprule
  test error & $\sigma$ \\
  \midrule
  0.1376 & 0.00087\\
\bottomrule
\end{tabular}\caption{Results copied from \citep{Dvovrak2007} which used a decision tree with softening splits.  In \citep{Dvovrak2007} the original
data set was split randomly into 7 different test cases.  Here those results are averaged and the standard deviation computed.  Result indicate that results using a the discontinuous polynomial neural network are competitive for $N_{p}>3$ when compared with Table \ref{T:GammaRay}.
$\sigma$ is the standard deviation of the results.}\label{T:GammaRay2}
\end{center}
\end{table}

\clearpage
\subsection{Handwritten Digit Recognition}\label{S:MNIST}
\begin{figure}
\begin{center}
\includegraphics[scale=0.6]{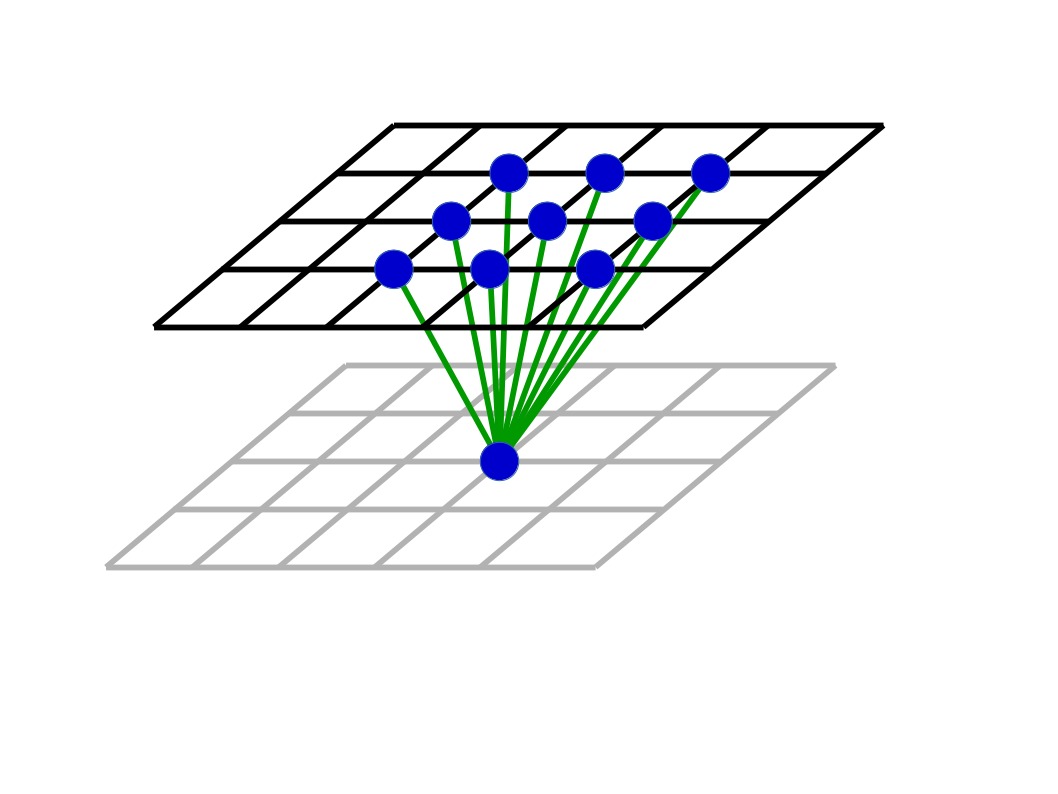}
\caption{Nearest neighbor connectivity used on the MNIST problem.  In this case the unit in the lower layer is
connected to the N nearest neighbors in the previous layer.  In the above example there are 9 nearest neighbors and the \textbf{width} of the stencil is defined to be 1.  A stencil with \textbf{width}=2 would have 25 nearest neighbors.}\label{F:NearestNeighbors}
\end{center}
\end{figure}

The MNIST is a standard benchmark of neural network codes on optical character recognition \citep{Lecun1998}.  The MNIST data set consists of a 60,000 image training set of 10 digits written from NIST employees and high schoolers.  In addition there is a 10,000 image test set that is used to test the generalization capability of the network after the network is trained on the training set. During parameter scans the test set is randomly split into test and validation groups with 50000 in the test groups and 10000 examples in the validation groups.  The images consist of 28X28 pixel 1 byte/pixel gray scale images.  As such, there are 728 input links (one for each input pixel) and 728 output links.  3 hidden layer are used in the result that are presented with a width of 7 as defined in \ref{F:NearestNeighbors}.  The digit recognition problem is solved using an autoencoder for each digit.  There are 10 autoencoders trained with each of the 10 digits.  The  Autoencoder 0 is only trained on the digit 0 examples, autoencoder 1 is only trained on the digit 1 examples etc...  At the end, the test set examples are run through each autoencoder, the predicted digit is determined by the autoencoder that produces the smallest error for the given input.  The error used in this paper is reconstruction error, improvements in this error measure for this approach to classification are explored in \citep{Kamyshanska2013}.  

Parameter scans were performed by varying the number of sub links from 1 to 14.  The width was varied from 2 to 7 and $N_{p}$ was varied from 2 to 6.  The best configuration validation results are presented in Table \ref{T:MNIST1} with results based on the results of 10 different validation and test sets based on shuffling of the 60,000 example complete test set.  

\begin{itemize}
\item $N_{p}$ - the number of Chebyshev-Lobatto points in the interpolation.
\item layers - the number of neurons in each hidden layer.
\item sub links - the number of sub links in each link.
\item width - the width of the stencil as defined in Figure \ref{F:NearestNeighbors}.
\end{itemize}

\begin{table}
\begin{center}
\begin{tabular}{cccc}
\toprule
  $N_{p}$ & sub links & validation error & $\sigma$ \\
  \midrule
  2 & 12 & 0.07124 & 0.00276\\
  3 & 10 & 0.05063 & 0.00293\\
  4 & 10 & 0.04804 & 0.00216\\
  5 & 6 & 0.04762 & 0.00271\\
  6 & 6 & 0.04879 & 0.00295\\
  \bottomrule
\end{tabular}\caption{Autoencoder MNIST validation results running for 1 epoch based on the best network parameters for the given $N_{p}$.  Note that 1 epoch is equivalent to presenting 5000 examples for each of the autoencoders.  The 3 point interpolation performs significantly
better than the 2 point interpolation, and the performance increases to $N_{p}=5$.  The best average performance is achieved for $N_{p}=5$.}\label{T:MNIST1}
\end{center}
\end{table}

\begin{table}
\begin{center}
\begin{tabular}{ccccc}
\toprule
  $N_{p}$ & 2 & 3 & 4 & 5 \\
  \midrule
  3 & 1.5e-9\\
  4 & 2.3e-11 & 6.1\\
  5 & 8.6e-11 & 4.7 & 100\\
  6 & 4.0e-10 & 29 & 83 & 60\\
\bottomrule
\end{tabular}
\caption{Percent risk that the means are actually equal in Table\ref{T:MNIST1} for each $N_{p}$ which is a measure of the significance of the difference of the means.  This table shows that $N_{p}=6$ is not significantly different than $N_{p}$=3,4,5 using 5\% risk.  $N_{p}$=5 however is significantly better than $N_{p}$=2,3 assuming a risk of 5\%}
\label{T:MNISTAccuracy}
\end{center}
\end{table}

Table \ref{T:MNIST1} shows MNIST results for $N_{p}=2 \text{ to } 6$ varying the number of neighbors used in the multi-layer autoencoder.  At a width of 7, a signal from the upper left corner of the input image is able to interact with the signal from the lower right.  T-test were performed to measure the significance of the difference in the measured means, see Table \ref{T:MNISTAccuracy}.  In this particular problem it was shown that all cases with $N_{p}>2$ were much better than the case with $N_{p}=2$.  In addition $N_{p}=4$ and $N_{p}=5$ was significantly better than $N_{p}=3$ assuming a risk of 7\%.  However, the difference in $N_{p}=4$ and $N_{p}=5$ is statistically insignificant.  Also, note that $N_{p}=6$ is not significantly better than $N_{p}=3$.

Recall, that for larger $N_{p}$ not only does the order of accuracy increase, but also the amount of compression (see Table \ref{T:OverShoot}).  This means that although the order of accuracy is increasing the
network inputs may be being compressed to a smaller region of the polynomial, and therefore reducing the effectiveness of increasing the order of accuracy.  In Table \ref{T:MNIST1} the number of sub links is always an even number, this is because an odd number of sub links typically gives a worse solution than an even number.  It's thought that there are 2 causes for this. (1) In deep networks the initial weight values are random values about 0 and the output of each link is averaged at the unit which tends to focus that output around the value 0.  As the number of layers is increased the output of each unit approaches 0 more closely.  If an odd number of sub links are used, the link function is smooth at 0.  If an even number of sub links are used then there is a discontinuity at 0.  The discontinuity near the origin means that the solution can rapidly jump between different sub networks and prevents the much slower convergence that is observed with an odd number of sub links.  As such, it's important that a discontinuity should occur at the origin.

\begin{table}
\begin{center}
\begin{tabular}{cccc}
\toprule
  $N_{p}$ & sub links & training error & test error \\
  \midrule
  5 & 6 & 0.007183 & 0.0145\\
  \bottomrule
\end{tabular}\caption{Autoencoder MNIST test results running for 50 epochs based on the best network parameters in Table \ref{T:MNIST1}.  In this case the full training set was used for training.}\label{T:MNISTFinal}
\end{center}
\end{table}
Table \ref{T:MNISTFinal} shows the best case from the validation tests in Table \ref{T:MNIST1} run out to 50 epochs.  Here we achieve a test error rate of 1.45\%, which can be compared with \citep{Kamyshanska2013} and \citep{Irsoy2014} which show results of 1.27\% and
1.1\% respectively.  Despite the performance discrepancy on this problem we do believe that there are a huge number of options for improving performance.  This simplest thing to do would be to add in the non-linearity at the neuron.  In this case, to keep the polynomial nature of the network, we suggest using the rectified linear unit.  In addition, as the network approaches convergence the network for a given input settles on a single sub network.  In this case, that network can be described by a smooth polynomial.  This suggest that a non-convex optimization technique should be used (even for the case of 0 discontinuities) and one might be able to take advantage of techniques specifically designed for high order polynomials such as \citep{Lasserre2001} or \citep{Laurent2009}.  Techniques for improving the performance are left as future work.
\clearpage
\subsection{Timing Results}
A critical question is how much computational time is added for $N_{p}>2$.  Both the derivatives and function evaluations become more complex as $N_{p}$ increases.  Furthermore, it might seem that adding additional sub links should increase the computational time.  Theoretically, adding new sub links does not increase the time for a single iteration.  With evenly spaced sub links the active sub link is determined in O(1) time, so it's independent of the number of sub links.  However, adding new sub links does change the memory usage and structure and can increase inefficiencies that way.

In Tables \ref{T:TimingSameLinks} and \ref{T:TimingSameDOF} below we run
the MNIST problem as above with 5 layers, and width 6 with the other parameters specified in the table.  The results were computed by running the test case 10 times and averaging the main loop time.  Adding additional sub links is not cost free, $N_{p}=2$ with 6 sub links is 12\% slower than $N_{p}=2$ with 2 sub links, but with 3 times the degrees of freedom - this is despite the fact the only one sub link is ever active.  In addition increasing the number of sub links with $N_{p}=2$ to 12 and the time jumps to 111 seconds (122\% slower than with 2 sub links) for 1000 iterations.  Despite this fact, the user still gets greater computational power without significantly increasing computational time.  Table \ref{T:TimingSameLinks} shows
that $N_{p}=6$ is only 2.2 times slower than $N_{p}=2$ despite having 3 times the degrees of freedom.  Similarly, in Table \ref{T:TimingSameDOF}
with the same number of degrees of freedom $N_{p}=6$ is only 1.83 times slower on average.

% Actual data
%5	1
%6.1	1.22
%7.7	1.54
%9	1.8
%10.8	2.16

\begin{table}
\begin{center}
\begin{tabular}{ cccc }
\toprule
$N_{p}$ & sub links & $t_{N_{p}}$ (seconds) & $t_{N_{p}}/t_{N_{2}}$\\
\midrule
2 & 2 & 50 &	1 \\
3 & 2 & 61 & 1.2 \\
4 & 2 & 77 & 1.5 \\
5 & 2 & 90 &	1.8 \\
6 & 2 & 108 & 2.2 \\
\bottomrule
\end{tabular}\caption{Timing results for equal number of links for 1000 inputs.}\label{T:TimingSameLinks}
\end{center}
\end{table}  

%2 & 6 &  5.914 \\
%3 & 4 &  6.5138 \\
%4 & 3 &  8.9975\\
%6 & 2 &  10.797 \\

\begin{table}
\begin{center}
\begin{tabular}{ cccc }
\toprule
 $N_{p}$ & sub links & $t_{N_{p}}$ (seconds) & $t_{N_{p}}/t_{N_{2}}$\\
 \midrule
2 & 6 & 59 	& 1.0 \\
3 & 4 & 65 	& 1.1\\
4 & 3 & 90 	& 1.5\\
6 & 2 & 108 	& 1.8\\
\bottomrule
\end{tabular}\caption{Timing results for equal number of degrees of freedom for 1000 inputs.}\label{T:TimingSameDOF}
\end{center}
\end{table}
\clearpage
\section{Conclusion}\label{S:Conclusion}
A novel approach to artificial neural networks is described where the traditional neuronal non-linearity is eliminated in favor of a discontinuous piecewise polynomial discretization of the weights space of each link.  The use of discontinuous piecewise polynomial approximations leads to a network, which is the superposition of multiple  networks with a set of shared weights as only a subset of the total network is active for each input signal.  Standard backpropagation is used for error correction with the modification that sub links that do not fire are not included in the backpropagation step.  The dropout technique \citep{Hinton2012} is used to minimize over fitting.  It is found that piecewise quadratic polynomials generally produce much better results than piecewise linear for the same number of degrees of freedom and that moving to increasingly higher order polynomials can provide additional improvement.  We have successfully demonstrated good solutions to the MAGIC and MNIST data sets and expect more complicated problems can be solved as well using this algorithm.  Future work will include the addition of a neuronal non-linearity as well as investigation of non-convex optimization techniques.
%\begin{thebibliography}
%\bibliography{bibtex_database}  
%\bibliographystyle{plain}
%\end{thebibliography}

\end{document}